\def\year{2019}\relax
\documentclass[letterpaper]{article} 
\usepackage{aaai19}  
\usepackage{times}  
\usepackage{helvet}  
\usepackage{courier}  
\usepackage{url}  
\usepackage{graphicx}  
\usepackage{graphics}
\usepackage{booktabs}
\usepackage{floatrow}
\usepackage{amstext}
\usepackage{amsmath}
\begin{document}
%
\title{A Trustworthy, Responsible and Interpretable System\\ to Handle Chit Chat in Conversational Bots}
\author{
  Parag Agrawal \\
  Microsoft\\
  India \\
  \texttt{paragag@microsoft.com} \\
   \And
   Anshuman Suri \\
   Microsoft \\
   India \\
   \texttt{ansuri@microsoft.com} \\
   \And
   Tulasi Menon \\
   Microsoft \\
   India \\
   \texttt{tulasim@microsoft.com} \\
}
\maketitle

\begin{abstract}
Most often, chat-bots are built to solve the purpose of a search engine or a human assistant: Their primary goal is to provide information to the user or help them complete a task. However, these chat-bots are incapable of responding to unscripted queries like ``Hi, what's up", ``What's your favourite food". Human evaluation judgments show that 4 humans come to a consensus on the intent of a given query which is from chat domain only 77\% of the time, thus making it evident how non-trivial this task is. In our work, we show why it is difficult to break the chitchat space into clearly defined intents. We propose a system to handle this task in chat-bots, keeping in mind scalability, interpretability, appropriateness, trustworthiness, relevance and coverage. Our work introduces a pipeline for query understanding in chitchat using hierarchical intents as well as a way to use seq-seq auto-generation models in professional bots. We explore an interpretable model for chat domain detection and also show how various components such as adult/offensive classification, grammars/regex patterns, curated personality based responses, generic guided evasive responses and response generation models can be combined in a scalable way to solve this problem.
\end{abstract}

\section{Introduction}
\label{Introduction}
Fueled by advances in the field of natural language processing, service and product websites are deploying chat-bots to help users navigate through their offerings and answering basic questions. These bots are either rule-based or designed to yield responses in a general and straightforward tone, which can get dull and monotonous at times. Any initiative by users to engage in a conversation with the bot often ends up in frustration: an abundance of ``I don't know" responses can get exasperating. To make bots sound less bot-like and more engaging, a system to support chitchat is pivotal.

There has been a lot of research on handling task-based queries~\cite{li2017end}, retrieving answers for natural language queries from structured~\cite{tablan2008natural} and unstructured documents~\cite{yan2016docchat}, and answering real-time information based queries like finance and weather~\cite{zhang2016joint}. We explore a system for detecting and answering chit-chat based queries. Training a sequence-to-sequence model is quite common to solve natural language problems. However, there are a few issues with such a model:

\begin{itemize}
    \item Generated responses can be risky for professional bots. A bot replying to a query like ``Hey, do you think Hitler was cool?" with ``Yes, I do" can be disastrous.
    \item These models either require computationally expensive GPUs to run at inference or are not fast enough for deployment in real-time systems.
    \item The lack of chat data available makes it infeasible to train a deep generative model. Research work often uses public Twitter~\cite{li2015diversity} and Reddit~\cite{willemsen2017sequence} data to overcome this problem: using conversations on threads as a proxy for chat. However, this data is not representative of private chat, which is much more personal and involved.
\end{itemize}

\begin{table*}[t]\centering
  \caption{Sample Intents, along with some queries which map to these intents.}
  \label{SampleIntentsTable}
  \centering
    \begin{tabular}{lll}
    \toprule
    \cmidrule(r){1-2}
    FriendlyName & Sample Queries & Type \\
    \midrule
    Greetings\textunderscore Generic & Hey happy weekends & Generic\\&Hi, hope you are enjoying your day \\ \cline{2-3}
    Greetings\textunderscore Bye & Bye Bye  & Specific\\ & See you.  \\\cline{2-3}
    Greetings\textunderscore GoodMorning & Good Morning & Specific\\ & Have a good morning  \\\cline{1-3}
    Bot\textunderscore Opinion\textunderscore Generic & What do you think about Trump? & Generic\\& What's your opinion regarding Hollywood movies?\\\cline{2-3}
    Bot\textunderscore Opinion\textunderscore Love & What do you think about love? & Specific\\\cline{2-3}
    Bot\textunderscore Opinion\textunderscore UserLooks & How do I look? & Specific\\
    & Am I looking pretty?\\\cline{2-3}
    \bottomrule
  \end{tabular}
\end{table*}

In our work, we define intents for chit-chat and design a model for intent-detection in user queries. Defining classes for intents is subjective: there can be many bases to determine intents. For example, intents can be segregated by sentiment, object and subject of the question, the content of a question, or based on the type of query (statement, command, question). Intents can be overlapping as well as strict subsets of each other as evident in Table \ref{SampleIntentsTable}. For instance, the intent \textit{GreetingsGeneric} (Hi, Hello, Bye, Good Night) is supposed to cover all kinds of greetings, but specific intents such as \textit{GreetingsGoodMorning} are also crafted, given the high volume of these queries. There also exists a \textit{Generic Intent} for \textit{UserStatement} which overlaps with greetings, since greetings are also an understatement. Thus, even for humans, intent classification can be quite ambiguous. Our system tries to model the chitchat space in such a way that that the responses can remain trustworthy and relevant while being interesting. It also provides a signal to seq2seq auto-generation model to ensure that it only generates a response for queries which are safe.

\section{Related Work}
Task and search intents in bots or personal assistants have been widely studied. The advantages of bots such as discoverability, availability and contextual understanding are often discussed in literature~\cite{Klopfenstein:2017:RBS:3064663.3064672}. \citeauthor{Klopfenstein:2017:RBS:3064663.3064672} discuss bots that are being used as functional replacements of mobile applications and name them ``Botplications". They also point out the ease with which a bot can be trained, using platforms like IBM's Watson\footnote{https://www.ibm.com/watson/} for question-answering. However, they do not discuss the need for a platform to train bots that can understand usual chit-chat.

In their recent work, \citeauthor{DBLP:journals/corr/AkasakiK17} show how bots need to chat and engage users apart from being task-oriented~\cite{DBLP:journals/corr/AkasakiK17}. They also show the complexity of natural language understanding in chat domain and why it cannot be treated as another domain determination problem.
Although joint intent (or domain) determination and slot-filling has been widely studied to improve accuracy, the same approach is not feasible in chat detection. Chat domain detection in spite of being related to intent and domain determination and heavily studied in the field of Task-Oriented Dialog System ~\cite{zhang2016joint,guo2014joint,tang2014learning} is significantly different from it. We propose a solution for not only chat-domain detection but extend it further to defining and classifying it into sub-intents.

Non-task and non-information oriented dialogue systems have been well studied~\cite{vinyals2015neural}. \citeauthor{sutskever2014sequence} explore a sequence-to-sequence model to generate responses dynamically ~\cite{sutskever2014sequence}, which is later extended in work by \citeauthor{sordoni2015neural}~\cite{sordoni2015neural}. Further, \citeauthor{DBLP:journals/corr/LiGBGD16} extend the sequence-to-sequence model to give responses based on a specific personality~\cite{DBLP:journals/corr/LiGBGD16}. These can't be used as stand-alone systems for the Chit-Chat domain, as discussed in Section~\ref{Introduction}.

\section{Our Approach}

\begin{figure}[!t]
\centering
\includegraphics[width=0.9\linewidth]{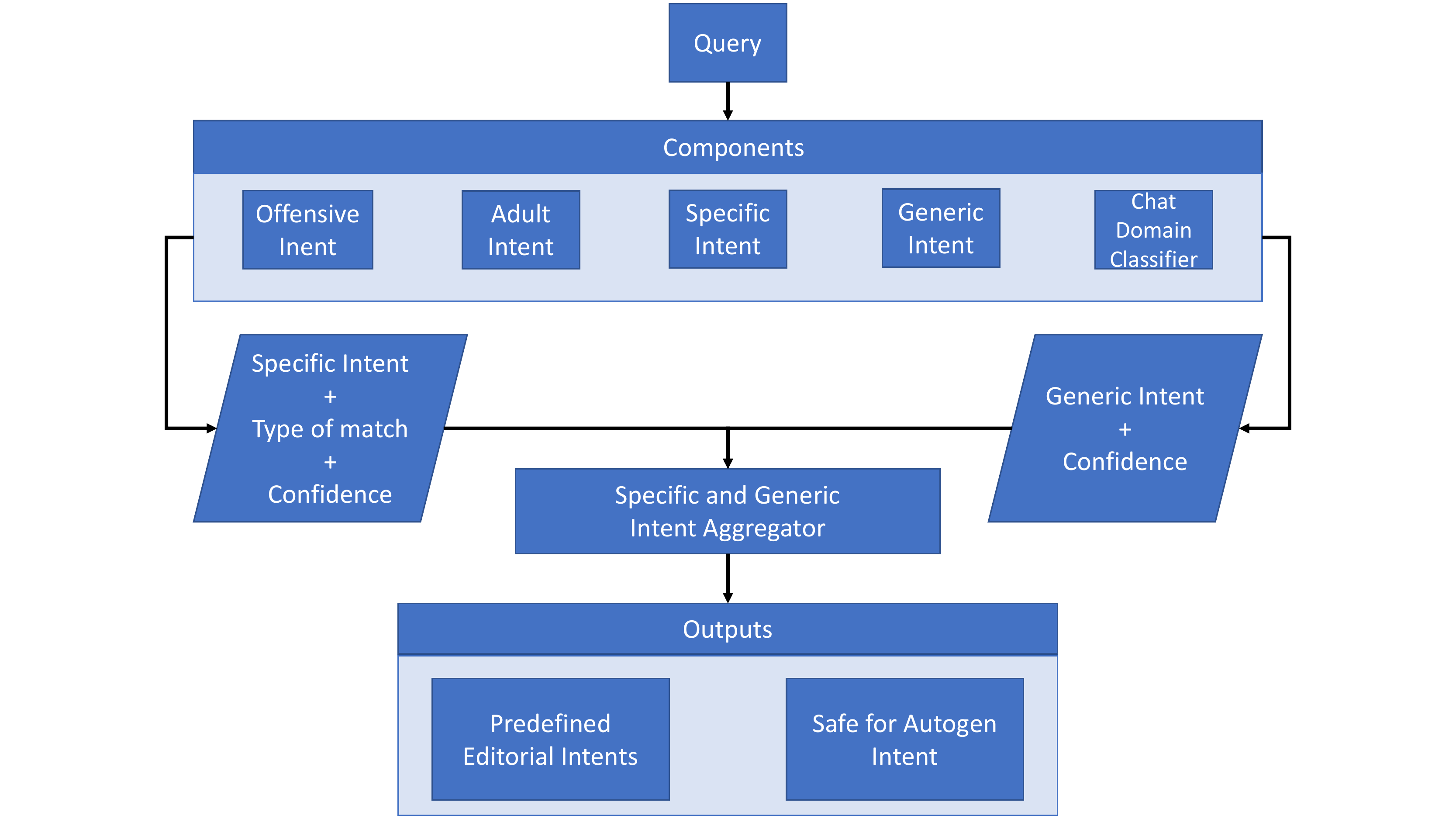}
\caption{\small Component-level flow of the proposed query understanding system for chat}
\label{architecture}
\end{figure}

Figure \ref{architecture} shows the proposed query understanding system for chat queries. In the following sections, we cover each of these blocks; explaining their importance and how they interweave with each other. Our system takes as input a user query and decides if it is a chat domain query and can be mapped to a predefined intent. It also determines if it's safe for a sequence-to-sequence auto-generation model to answer. For instance, queries such as ``What do you think about me", ``Do you think I should die?", ``Do you think \{Politician\} is corrupt" should not be answered by a generative model, since its responses are not curated and can give answers which might hurt the user's sentiments. Our current system uses more than 100 predefined intents, out of which 20 are generic. 

\subsection{Chat Domain classifier}

Segregating chat queries from goal/information oriented queries is crucial for a chat bot~\cite{DBLP:journals/corr/AkasakiK17}. We build a classifier which models the probability of a given query belonging to chit-chat domain. Obtaining high-quality training data for such a task is cumbersome since this field has been little explored. To address this lack of data, we got distinct queries judged by a team of human annotators. In addition to this dataset, we used data from other domains. We experimented with a vast array of methods to train this domain classifier. Discussing all of them here is out of the scope of this paper. The best results that we obtained were using an ensemble of models explained in Section \ref{ChatDomainFeatureModelling} which treat lexical and semantic features differently. 

\subsubsection{Data Preparation}
For this problem, we curated a good amount of distinct queries made to a popular personal-assistant chat-bot. After performing data anonymization, we got it judged via crowd-sourcing. The annotation was done using four labels: \textit{CHAT}, \textit{TASK}, \textit{INFORMATION}, and \textit{JUNK}. Four judges labelled each query. We considered \textit{TASK}, \textit{INFORMATION} and \textit{JUNK} as not being chat, but it might depend on the type of bot the classifier is being trained for. A fuzzy, high-recall bot might consider \textit{JUNK} as Chat-Domain while training the classifier. Only the queries where at least 3 out of 4 judges marked it as \textit{CHAT} were considered to be positive samples. We ignored queries where two or fewer judges marked a query as \textit{CHAT} using the others as negative samples, which yielded a distribution of $\sim$19\% positive samples, $\sim$75\% negative samples, and 6\% ignored samples.

\begin{figure}
\includegraphics[scale=0.3]{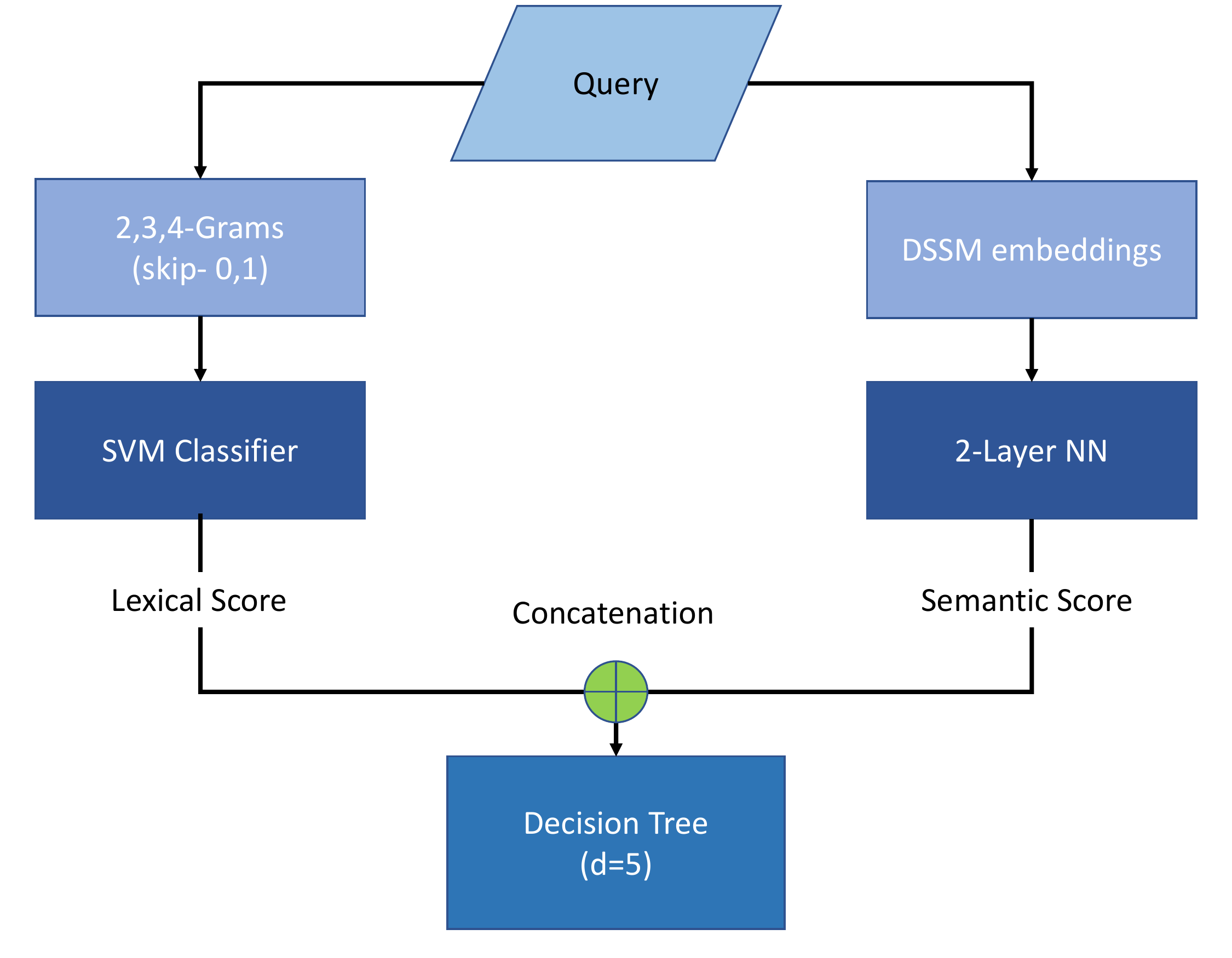}
\caption{Chit-Chat Domain classifier}
\label{Chat Domain Classifier}
\end{figure}
\begin{table}
  \begin{tabular}{lll}
    \toprule
    \cmidrule(r){1-2}
    Lexical Features & AUC & Accuracy \\
    \midrule
    1,2,3 word gram & 85\%  & 83\% \\
    (TF-IDF) & & \\
    \hline
    1,2,3 word gram & 88\%  & 87\% \\
    (TF-IDF) & & \\
    and 3-char-gram (TF) & & \\
    \hline
    1,2,3,4 and 1-skip  & 14\%  & 12\%  \\
    word-gram (TF-IDF) & & \\
     and 3-char-gram (TF) & & \\
    \bottomrule
  \end{tabular}
   \caption{Comparison using different lexical features}
  \label{ChatDomainClassifierLexicalTable}
\end{table}

To \textbf{augment} this data, curated and high-impression queries from domains such as Weather, Finance, Maps, QnA, Tasks (Reminder, Alarms) are taken as \textbf{negative samples}. Such queries are examples of cases where the user's utterance does not have a chit-chat intent since they want to explore specific services.

For some queries, the search-engine was triggered frequently. At the same time, these users were not satisfied with the results: this is probably because of a user trying to initiate a conversation, but being misinterpreted by the bot as a search query. Such cases were, thus, taken into consideration as \textbf{positive samples}. We infer dissatisfaction by the click-through rate for queries to the search engine as well as similar re-queries soon after. \\
It may be noted that this semi-supervised approach of data augmentation can lead to noisy training data. However, all of the queries that we consider in the method above are high volume queries, thus highly unlikely to contain a substantial amount of noise.

After post-processing and cleaning of data, 0.03 million positive samples and 0.3 million negative samples were obtained. Instances from judged samples were given a higher weight (5:1) than augmented data since human judgments are much more strict and reliable.

\subsubsection{Features and Modelling}
\label{ChatDomainFeatureModelling}
We represent any given query as a combination of lexical and semantic features. The semantic component includes a 300-dimensional vector representation of the query obtained using a  pre-trained DSSM-source model \cite{a-latent-semantic-model-with-convolutional-pooling-structure-for-information-retrieval}. Lexical features from the query are extracted as N-Grams: 1,2,3,4 word grams with one skip and also 3-char grams. Performance for different combinations of these lexical features is shown in Table~\ref{ChatDomainClassifierLexicalTable}.

As shown in Figure \ref{Chat Domain Classifier}, we trained an ensemble of models, where a Support Vector Machine (SVM)~\cite{cortes1995support} is used for lexical features, and a two-layer neural net is used for semantic features. A decision tree with a maximum depth of 5 is used to combine these lexical and semantic scores. Our final weighted Precision, Recall and F1 was 94.9\%, 80.2\%, and 86.93\% respectively, which is significantly better than the approach described by~\citeauthor{DBLP:journals/corr/AkasakiK17}~\cite{DBLP:journals/corr/AkasakiK17} though that might be due to the difference in data set. We used a much bigger dataset for training and measured weighted Precision|Recall whereas \citeauthor{DBLP:journals/corr/AkasakiK17} uses a randomly sampled set from 3 sets of queries based on frequency.

Decoupling lexical and semantic models is motivated by integrating \textbf{interpetability} in our system. Possessing the ability to interpret a model makes understanding its predictions easier and helps provide a strong intuition behind the weights it learns. Interpretability is a much-needed property that is often missing from deep models, owing to the humongous number of parameters and non-linearities in them. A linear model with N-Gram features gives the flexibility to tune the model as per the one's needs. For instance, if user feedback indicates that a model is incorrectly detecting queries like ``Is it raining in {location}" as being chit-chat query, the 3 gram of ``raining $\mid$ in $\mid$ \{location\}" can be given a negative weight to tackle these false negatives.

\subsection{Offensive and Adult classifiers}
\label{Offensive and Adult classifiers}

 Detection of inappropriate queries in conversation is a relevant problem and has been well explored in literature~\cite{yenala2017deep}. We used the Content Moderation API\footnote{https://azure.microsoft.com/en-us/services/cognitive-services/content-moderator/} from Microsoft Cognitive Services to identify Obscene and Offensive text. The generic intent component later consumes this signal as a feature to the multiclass classifier. We have generic intents like \textit{criticism\textunderscore abusive} defined for inappropriate queries.

\subsection{Specific Intent}

\begin{figure}[!t]
\centering
\includegraphics[width=0.9\linewidth]{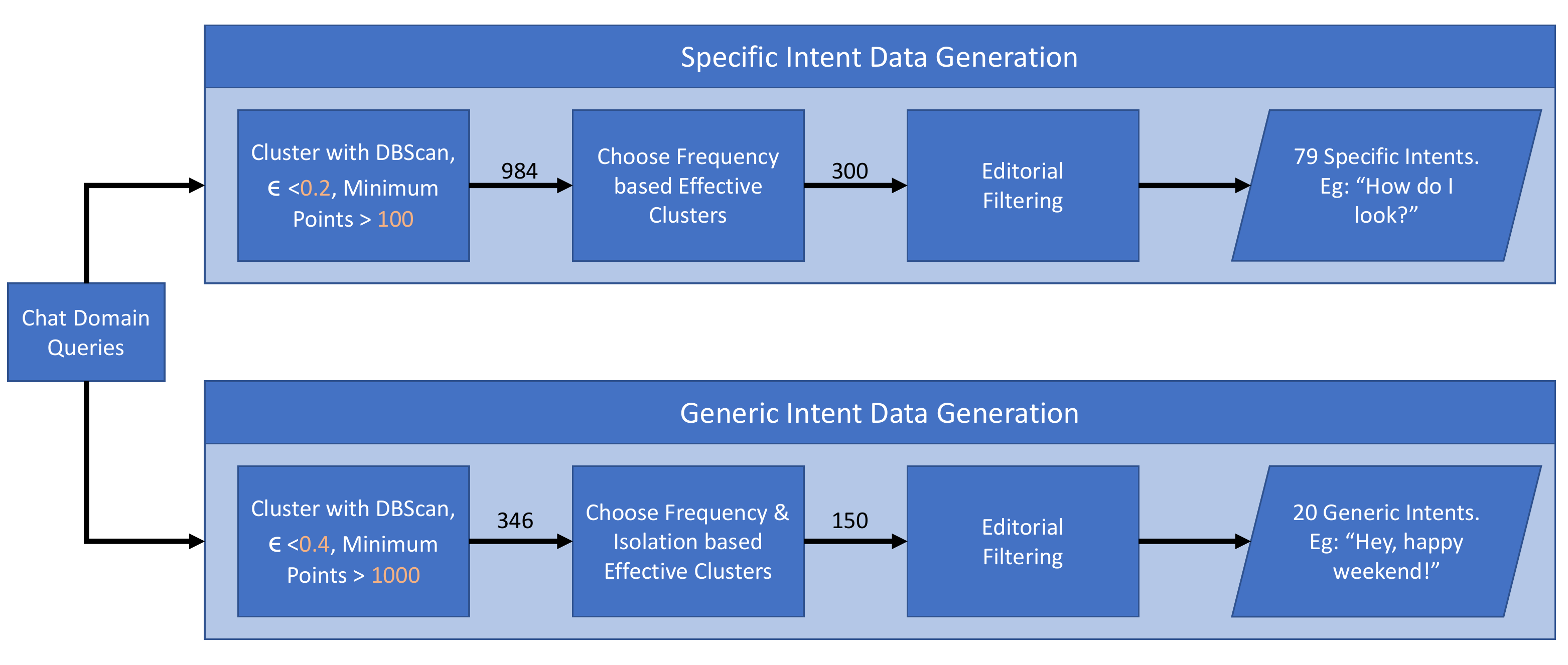}
\caption{\small Offline flow for defining specific and generic intents}
\label{DefiningIntents}
\end{figure}

\textit{Specific Intents} are intents which are very clear and directed. For example, ``Tell me a joke" is a \textit{Specific Intent}: the user is specifically looking for a joke. \textit{Specific Intent} is something which the bot can clearly understand and is capable of answering it directly. As expected, it is infeasible to consider all possible cases for \textit{Specific Intents}. To optimise for coverage while keeping the number of specific intents reasonable, intents which have the highest amount of query impressions are defined to be \textit{Specific Intents}.

\subsubsection{Defining Specific Intents}
\label{DefiningSpecificIntents}
User chat logs were clustered using DBSCAN \cite{ester1996density}, using cosine distance between DSSM-source embeddings of queries as the distance metric. While clustering, the maximum radius ($\epsilon$) was fixed at 0.2, and the minimum number of distinct queries as 100. A low radius is picked because we only want queries which are very similar to be clustered together. Clustering was done on distinct chat domain queries, giving us a total of 984 clusters.
For shortlisting effective clusters out of these, we define an effectiveness factor which takes into consideration the frequency of various distinct queries.
\newline

Frequency based Effectiveness of $i^{th}$ \textit{Specific Cluster} =
\begin{equation} \label{eq:effectiveness}
\sum_{i=1}^{n} (1-D_i)*W_i
\end{equation}

where $D_i$ is the distance of the instance \textit{i} from the centroid of its cluster, and $W_i$ is the weight (number of impressions count) of that instance.

We define effectiveness to optimise for the number of specific intents: minimising the amount of human annotation required while maximising keeping coverage over the query-sample space. Using this effectiveness criterion, the top 300 clusters are sent to be filtered and edited by a trained content-writing team that has a fair amount of experience in the chit-chat domain. The possible operations were:
\begin{itemize}
    \item rejecting a cluster with reason,
    \item choosing a cluster as intent, and
    \item merging one cluster to other chosen cluster.
\end{itemize}

Out of all clusters, we rank and pick the top 300 for human evaluation. 79 out of these clusters are marked as \textit{Specific Intent}. After this step, 132 clusters are merged into these 79 clusters. 114 clusters are rejected because of being ambiguous/bad clusters, while the remaining 10 have non-chat queries present.

As indicated by these statistics, the clustering yielded a considerable amount of ambiguous clusters with no clear intent. As discussed in Section \ref{Introduction}, it's a hard problem to cluster these queries into readily perceivable intents.  Since the query set is first passed through Chat Domain Classifier which filters out non-chat queries with a precision of $\sim$91\% (Section~\ref{ChatDomainFeatureModelling}), there are few non-chat clusters. However, this approach yields the required number of \textit{Specific Intents}. The purpose behind having \textit{Specific Intent} is to have distinctive and declarative responses, which can be useful to define the personality of a bot. Thus, precision for \textit{Specific Intent} matching needs to be very high. For instance, if there's an intent for ``Do you like ice-cream?" to which the response might be ``Yes, I love it", we don't want it to fire for ``Do you like ice-cream when it's not frozen?", as this may be inconsistent with the bot's personality.

\subsubsection{Specific-Intent Runtime}

\begin{figure}
\begin{floatrow}
\ffigbox{%
  \includegraphics[width=0.85\linewidth]{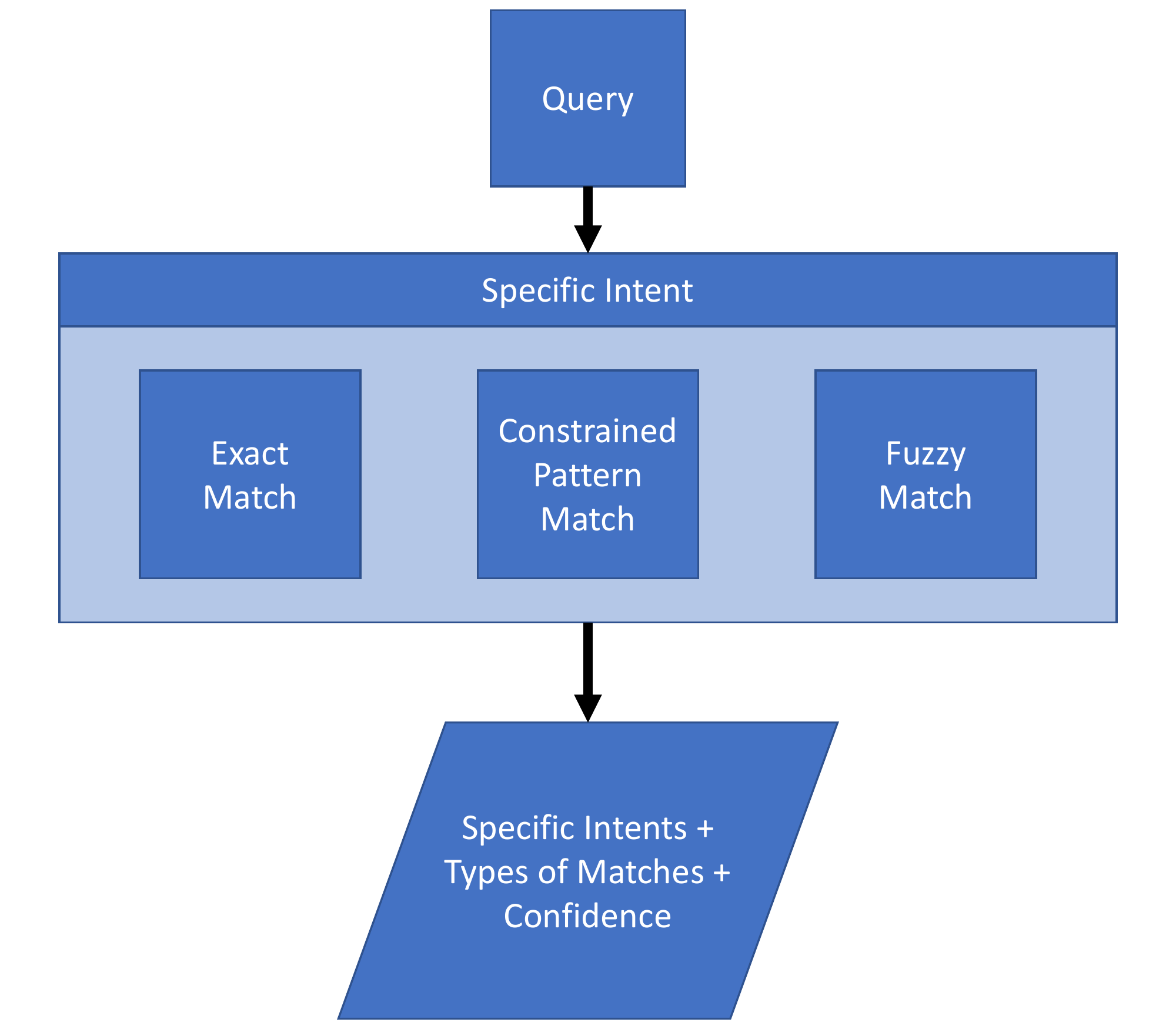}
}{%
  \caption{Runtime flow for classifying a query into a specific intent.}
    \label{SpecificIntentsRuntime}
}
\ffigbox{%
  \includegraphics[width=1\linewidth]{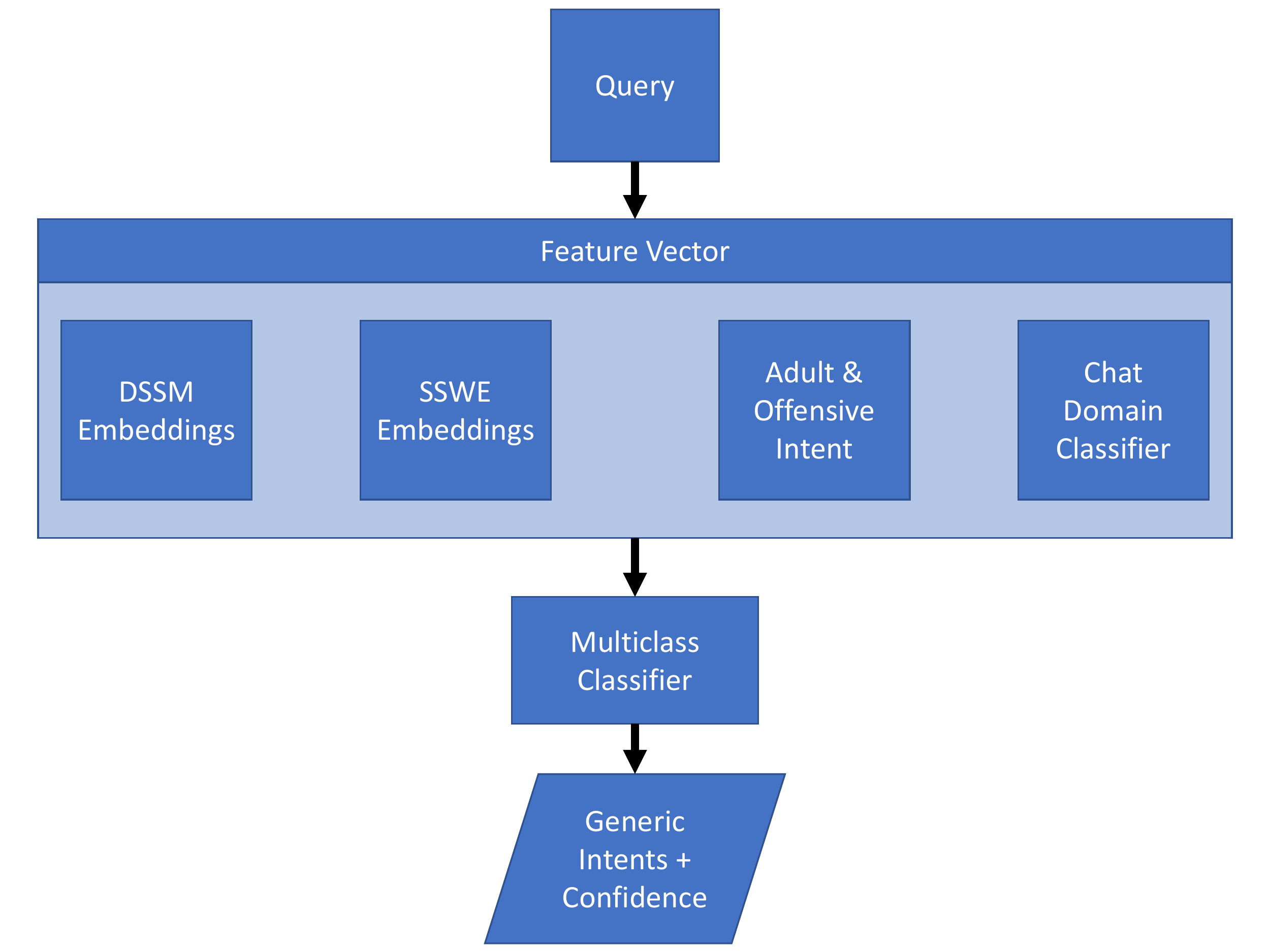}
}{%
   \caption{Runtime flow for classifying a query into a generic intent.}
    \label{GenericIntetntsRuntime}
}
\end{floatrow}
\end{figure}

For \textit{Specific Intent}, we consider exact matches, constrained pattern matches, and fuzzy match at a very high threshold.
Exact matches occur when the user's query is present in our query base for \textit{Specific Intents}. Only basic normalization such as case insensitivity and removing stop-words/junk-characters are handled while performing matching. Pattern matching is done on pre-mined or curated grammars or regexes.

This fuzzy match is performed using \textbf{cosine similarity} between pre-computed sentence DSSM-source embeddings and user query DSSM-source embeddings. The score for a specific intent is computed as:
\begin{eqnarray} \label{eq:cosine}
\max_{q\in [0, n)} \frac{\langle A[q], B \rangle}{\| A[q] \|_{2} \cdot \| B \|_{2}}
\end{eqnarray}
where $A[q]$ is a 300-dimensional embedding of $q^{th}$ for the given intent, and $B$ is a 300-dimensional vector representation of user's query. 

\textit{Specific Intents} with a score $\geq0.9$ are shortlisted for use. Setting this threshold to a high value helps handle spelling mistakes, synonyms and lemmatised representations of words.

\subsection{Generic Intent}
As discussed in the section above, \textit{Specific Intents} only cover a small portion of chat domain intents: it covers only the most frequent, leaving many gaps (Table \ref{SampleIntentsTable}). To fill these gaps, we propose \textit{Generic Intents}. 

\subsubsection{Defining Generic Intents}

As shown in Figure~\ref{DefiningIntents}, \textit{Generic Intents} are extracted using DBSCAN based clustering, similar to the method described in Section~\ref{DefiningSpecificIntents}. We set $\epsilon$ to a high value, \textit{i.e.} 0.4 to keep the clustering loose. At the same time, we set the minimum number of instances to 1000; a \textit{Generic Intent} should have a lot of distinct queries, to increase its cover. \\
To calculate the effectiveness of these clusters, we redefine our metric: the distance between neighbouring clusters is also taken into consideration. Since generic classes can easily form ambiguous clusters for a high radius, factoring in proximity to neighbouring clusters helps reward distinct and non-overlapping clusters with a better score. For example, the query ``What do you think about love" can be treated as a \textit{Generic Intent} of ``bot\textunderscore opinion\textunderscore generic", which represents the user asking a bot for its opinion about anything. For such intent, the cluster would contain queries like ``What do you think about ...", ``Do you think... is great". At the same time, the same query may be classified under ``love\textunderscore generic", where the user is talking about love. In which case, it will be clustered with queries such as ``I am in love", ``Tell me a love story". These kind of queries are ambiguous for both the machine as well as humans to judge.

Frequency and Isolation based Effectiveness of $i^{th}$ \textit{Generic Cluster} =

\begin{equation} \label{eq:generic_effectiveness}
DC_{min}^2 * \sum_{i=1}^{n} (1-D_i)*W_i
\end{equation}

where $D_{i}$ is the distance of the instance \textit{i} from the centroid of its cluster, and $DC_{min}$ is the distance of centroid of the cluster from its closest neighbouring cluster's centroid. \\
The top 150 clusters, ranked by this effectiveness score, were sent for human annotation (Section~\ref{DefiningSpecificIntents}).

Out of all clusters, we rank and pick the top 150 for editorial filtering. 20 clusters are chosen as \textit{Generic Intent} and 45 clusters are merged into these 20 clusters. 33 clusters are rejected because of being ambiguous (non-distinctive), while the remaining 52 are rejected because of the low volume of chat domain queries in them.

For generic intents like ``Compliment\textunderscore Bot" and ``Criticism\textunderscore Bot", a specific response cannot be provided, but a redirection/suggestion can be given. For instance, the bot may know that the user is complimenting it, but does not exactly know what the compliment is about. It could be looks (\textit{Specific Intents}: ``Compliment\textunderscore Looks") or humour (\textit{Specific Intents}: ``Compliment\textunderscore Humor"). Thus, \textit{Generic Intents} help prompt for suggestions or redirect towards specific intents which the bot can understand. Responses for \textit{Generic Intents} can be ``Did you mean to compliment me about my previous response" (a re-direction to a \textit{Specific Intents}) or ``If you like me, you can get more information here...", which is a suggestion. 

\subsubsection{\textit{GenericIntent} Runtime}

As shown in Figure~\ref{GenericIntetntsRuntime}, the multi-class classifier predicts the probability distribution over all generic classes, given a query. The features extracted from the query are:-
\begin{itemize}
    \item DSSM-source embeddings~\cite{a-latent-semantic-model-with-convolutional-pooling-structure-for-information-retrieval} for capturing the semantics of the sentence: \textbf{300 features}.
    \item SSWE (Sentiment Specific Word Embeddings)~\cite{tang2014learning} for capturing the sentiment of a query. DSSM captures the context of a sentence but fails to capture its sentiment. For instance, ``I love you" and ``I hate you" are opposites but are still close based on DSSM-source Embeddings. This is undesirable while classifying Chit-Chat intents: \textbf{150 features}.
    \item Signals from adult and offensive classifiers (Section~\ref{Offensive and Adult classifiers}): \textbf{2 features}.
    \item Signal for chat domain classifier (Section~\ref{Chat Domain Classifier}): \textbf{1 feature}.
\end{itemize}

\begin{table*}[t]\centering
  \caption{Component level measurements. $C_{unweighted}$, $C_{weighted}$ and $P_{unweighted}$ refer to unweighted coverage, weighted coverage and unweighted precision respectively.}
  \label{ComponentMeasurementTable}
  \centering
    \begin{tabular}{llll}
    \toprule
    \cmidrule(r){1-2}
    Component & $C_{unweighted}$ & $C_{unweighted}$ & $P_{unweighted}$ \\
    \midrule
    Specific Intent- Exact Match & 25\%  & 4\% & 98\% \\
    Specific Intent- Pattern Match & 17\%  & 9\% & 94\%\\
    Specific Intent- Fuzzy Match & 14\%  & 12\% & 91\% \\
    Generic Intent  & 43\%  & 75\% & 78\%    \\
    Overall  & 100\%  & 100\% & 86.76\%    \\
    \bottomrule
  \end{tabular}
\end{table*}

Concatenating all of the above features yields a feature vector with 453 dimensions. We train a fully connected neural network model with two hidden layers of 300 dimensions each, and Sigmoid activation after the hidden layer. The input and output layers have dimensions of 453 (number of features) and 20  (number of generic intents) respectively. This model is trained using queries from clustering which had at least 1000 queries per cluster. This criterion yielded 20 such classes.

\subsection{\textit{Generic} and \textit{Specific Intent} Aggregator}
 This component aggregates the results from \textit{Specific} and \textit{Generic Intent} to yield final intents, along with a signal denoting how safe is it for the auto-generation to respond.
Some of the rules for this aggregation component are:
\begin{itemize}
    \item If \textit{Generic Intent} of ``criticism\textunderscore generic" has a probability of greater than 0.5, \textit{safe\textunderscore for\textunderscore auto-generation} is set to false.
    \item If \textit{Specific Intent} occurs due to a fuzzy match, and a \textit{Generic Intent} of ``criticism\textunderscore response" is predicted with high confidence, the \textit{Specific Intent} is discarded. For instance, ``Don't tell me a joke" or ``That joke was really bad" could easily match with ``command\textunderscore joke", even though the user is criticizing the bot. 
\end{itemize}

After applying such rules, \textit{Generic Intents} are appended to \textit{Specific Intents}: whenever there is a \textit{Specific Intent} it will be given priority over \textit{Generic Intents}. For consistency, in such cases, scores from \textit{Generic Intent} are scaled to be lower than \textit{Specific Intents}.

\section{Results}

\begin{table*}[t]\centering
  \caption{Comparison with other chit-chat systems. $P_{unweighted}$ and $R_{unweighted}$ refer to unweighted precision and unweighted recall respectively.}
  \label{CompeteTable}
  \centering
    \begin{tabular}{lll}
    \toprule
    \cmidrule(r){1-2}
    Chat System &  $P_{unweighted}$ & $R_{unweighted}$ \\
    \midrule
    Using Exact Match for curated queries & 98\%  & 4\%\\
    \hline
    Well known Chit-Chat System (Threshold @ Best $F_{1}$) & 73\%  & 56\% \\
    \hline
    Our system (Threshold @ Best $F_{1}$) & 84\% & 75\%\\
    Our system with auto-generation (Threshold @ Best $F_{1}$) & 82\% & 79\% \\
    \bottomrule
  \end{tabular}
\end{table*}

We did measurements on the individual components shown in Figure 1. Results are documented in Table~\ref{ComponentMeasurementTable}. The \textit{Specific Intent}'s precision is greater than 90\% for exact and pattern matches, because it majorly depends on judged and hand-written queries/grammars. Fuzzy matching is also constrained using a tight threshold, giving a precision of 91\%. The weighted coverage (queries with frequency taken into consideration) for \textit{Specific Intent} is high. However, it can be observed that the un-weighted coverage (distinct queries) amounts to only 25\% (4 + 9 + 12) of the sample space, whereas \textit{Generic Intents} cover 75\% of the distinct query sample space. The precision of \textit{Generic Intent} is 78\%, owing to ambiguity and the complexity involved in chat domain. The responses for \textit{Generic Intents} are supposed to be evasive and suggestive: a precision of 78\% does not cause much dissatisfaction among users.

We evaluate our system on 5000 queries and compare it against an existing chat system in production for a complete analysis with existing chit-chat systems. A gap in recall denotes that the query was supposed to be a chat query but wasn't answered correctly, or wasn't answered at all. A gap in precision is indicative of a non-chat query being answered, or an incorrectly detected intent. Results are shown in Table~\ref{CompeteTable}. We focus on un-weighted (distinct queries) measurements, since weighted scores largely depend on the kind of bot in which the system is used, regular monitoring, and covering the head queries.

\section{Conclusion}

Our work aims to solve the problem of understanding intents in chat queries. Results show that each chat query cannot be represented as a specific intent; an intent-based \textbf{hierarchical approach} is much needed to solve this problem. We propose a system using various components that define sub-intents; partitioning the semantic chat space into explicit \textbf{perceivable intents} via clustering. The proposed method also aims at \textbf{reducing human effort} by choosing the most effective clusters for annotation. Our pipeline detects if a given user query is from the Chit-Chat domain, and maps it to appropriate intents. Thus, our system helps chat-bots \textbf{handle scenarios}, as well as provides a \textbf{reliable signal} for safely using auto-generation models.



\bibliographystyle{aaai}
\bibliography{aaai_19}

\end{document}